\documentclass{article}
\usepackage{spconf,amsmath,graphicx}
\usepackage{optidef, comment}
\usepackage{algorithm}
\usepackage{algorithmic}
\usepackage{caption}
\usepackage{lipsum}
\usepackage{subcaption}
\usepackage{adjustbox}
\usepackage{amsmath,amsfonts,amssymb,amsthm, bm}
\usepackage{mathtools, enumitem}
\usepackage{optidef}
\usepackage{tikz}
\usepackage{cite}
\usepackage{amsmath,amssymb,amsfonts}
\usepackage{algorithmic}
\usepackage{graphicx}
\usepackage{textcomp}
\usepackage{xcolor}
\usepackage{amsmath}
\usepackage{comment}
\usepackage{amssymb}
\usepackage{multirow}
\usepackage{multicol}
\usepackage{array}

\usepackage{blindtext}
\usepackage{xfrac}
\usepackage{amsmath}
\usepackage{graphicx}
\usepackage[spaces,hyphens]{url}
\usepackage{float}
\usepackage{multirow}
\usepackage{graphics}
\usepackage{subcaption, float, comment} 
\usepackage{tabularx}
\usepackage{multirow}
\usetikzlibrary{positioning}
\usepackage{mathtools, enumitem}

\DeclareMathAlphabet{\mathpzc}{OT1}{pzc}{m}{it}

\floatname{algorithm}{Algorithm}

\title{Fourier Double Layer Neural networks for Functional Approximation}
\name{R Subhash Chandra Bose, Kakarla Yaswanth}
\address{ Department of Electrical Engineering,\\Indian Institute of Technology Hyderabad \\Hyderabad, India}
\begin{document}
%
\maketitle
\begin{abstract}
The success of Neural networks in providing miraculous results when applied to a wide variety of tasks is astonishing. Insight in the working can be obtained by studying the universal approximation property of neural networks. It is proved extensively  that neural networks are universal approximators. Further it is proved that deep Neural networks are better approximators. It is specifically proved that for a narrow neural network to approximate a function which is otherwise implemented by a deep Neural network, the network take exponentially large number of neurons. In this work, we have implemented existing methodologies  for a variety of synthetic functions and identified their deficiencies.  Further, we examined that Fourier neural network is able to perform fairly good with only two layers in the neural network. A modified Fourier Neural network which has sinusoidal activation and two hidden layer is proposed and the results are tabulated.
 
\end{abstract}

\section{Introduction}
  Neural networks have been extensively popular these days. However the theory of why neural network work is still not well developed and extensive research is going on. Many explain the success of ANN by comparing the neuron in ANN that of a biological neural network in the brain.  

However, in our perspective, the miraculous success of ANNs can better be attributed to some other properties of ANNs like the "Universal property of Neural networks" which have thorough mathematical proofs underlying them. \cite{shamir2018resnets,lu2017expressive,eldan2016power,hanin2017approximating,telgarsky2016benefits,kidger2020universal}.
Universal Approximation  theorem states that the neural networks are able to approximate any function that connects inputs to outputs.

\section{Literature Survey}\label{literature survey}

\subsection{Approximating various classes of functions}
In this paper , we consider approximating various classes of functions, namely trigonometric function, polynomial function, exponential function, step wise increasing function etc. The target function is expressed in terms of taylor series expansions to large number of terms. All the terms in taylor series are  polynomial functions which are in turn implemented by suitable NN architectures\cite{liang2016deep}. 
Given a function f , let the neural network is simulating function g, the distance or error between functions is the maximum absolute difference over $ hyper cube [0,1]^d $

In the paper, the authors presented upper bounds for univariate starting with polynomial functions.

Two kinds of activation functions are used in this paper. ReLU and binary step units. The multiplication of two bits is implemented using a ReLU function. 
\subsubsection{Approximating function $x^2$}
In the method employed, the author wants to find the depth and size of neural network needed to approximate the function. First a simple function $x^2$ is considered. The quantity 'x' $\in (0,1)$.
Here, first 'x' is approximated as $\sum_0^{n} \frac{x_i}{2^i} .$
A n-layer neural network is discussed to find $x_i$'s. 
Next, the function $f^\sim (x) = f(\sum_0^{n} \frac{x_i}{2^i})$ is implemented by a two-layer neural network. To achieve $\epsilon-approximation$ error, n should be chosen as $n=\lceil log_2 \frac{1}{\epsilon} \rceil$+1. Such deep neural network has $\mathcal{O}(log\frac{1}{\epsilon})$ layers, $\mathcal{O}(log\frac{1}{\epsilon})$ binary step units and $\mathcal{O}(log\frac{1}{\epsilon})$ rectifier linear units.


In the Resnet paper, the functions are approximates to a concatenation of trapezoidal functions which is in turn implemented by Resnet.
\subsection{Feedforward Neural network as Function Approximator}
For this discussion, I took some of the concepts from 
\cite{liang2016deep}. Here, we discuss the question of function approximation using Neural network. The key intuition is that multiplying of two binary bits can be performed by ReLU. Similarly. two numbers can also be multiplied using ReLU. We will demonstrate these methods in this section.
\subsubsection{Binary representation of x using NN}
$Example:$ 
\begin{center}
$Let \ x_1 \ and \ x_2 \ be \ two\  bits\  to  \ be \ multiplied.
\newline
$\hspace{5mm} $ x_1 \in \{0,1\}\  and\  x_2 \in {0,1}. \ Let \ x_1*x_2=\ y. \newline
$So \ the\  Truth \ table\  for \ \textit{y} \ is\  as\  follows:


\begin{tabular}{ c c c }
 \texttt{$x_1$} & \texttt{$x_2$} & \texttt{$y$} \\ 
 0 & 0 & 0 \\  
 0 & 1 & 0\\ 
 1 & 0 & 0 \\  
 1 & 1 & 1    
\end{tabular}
\end{center}
This can be written as $y=max\{0,x_1+x_2-1\}$ \newline or more generally as $y=max\{0,k(x_1-1)+x_2\}$ where $k$ is any constant , such that $k \geq $1 .

Fig.1 gives the error between the function and taylor series approximation. The latter is taken for various number of polynomials  given by N. For example, N=5, indicated taylor series is calculated till degree 5 polynomial.
Various types of functions like sinusoidal, pollynomial, exponential functions are considered. 
It may appear not so as expected. According to Remainder theorem of Taylor series, for a smooth function one would expect the remainder which is correlated to epsilon1 to be very less. However we notice that the constant coeficient of x in the function f(x) decides the value of the remainder and for small N, this can be large as seen in the table for sin function.

Where as in case of epsilon2 error, its the  difference in value between the output of Neural network and the taylor series and Neural network output. Here the variable N corresponds to the number of terms in binary expansion.

• Here also the implementation is done in the same way as Resnet paper for various functions to get values for the errors $\epsilon_1$ and $\epsilon_2$
      
    • Here, For $\epsilon_1$ table, rows are represented by numbers of terms under consideration for Taylor series and columns are represented by functions.
      
    • And different tables are made for different points about which Taylor series is considered.
      
    • And a whole another set of tables are made keeping infinity norm under consideration.
      
      
    • For $\epsilon_2$ table, rows are represented by numbers of bits used and columns are represented by functions.
      
    • For $\epsilon_2$, there are zero errors for 60 and more bits.

\section{Preliminaries and Problem Statement}
We want to find an optimum Neural network architecture for a given class of functions. Different methods of constructing Neural networks to approximate the functions are discussed elaboratorately
Following the results of implementing the classes of functions in Srikant's paper and Resnet paper, following points are observed:
 1. Srikant's method is approximating polynomial functions better as expected.
 2. Resnet method better approximating step wise increasing / decreasing functions as expected.
 3. None of the methods approximate the sinusoidal functions satisfactorily.
 
 Hence, we propose a class of NN called fourier NN with sinusoidal activation function to deal with this problem.
 Our intuition is because the activation function is sinusoidal, it can easily approximate sinusoidal functions.
 Also since , by Fourier series , any function can be written in terms of sinusoidal functions. So, by using sinusoidal functions as building blocks, our Fourier Neural network can be used to construct other functions also.

In the literature, the optimality of the neural network architecture is not considered. In this paper, we address the problem that is it possible to optimise the number of layers or number of neurons required for the architecture to perform universal approximation? If so, how do we arrive at that architecture?  In this paper, a FNN is proposed to approximate sinusoidal class of functions. Further , we extend the FNN architechture to take two layer instead of single layer. This implements double trigonimetric functions affectively. Further, a skip connection is allowed from first later to the outputs allowing single trigonimetric functions as well. The trade of between width and depth is considered. It is proven that FNN with two layers is able to approximate better than single layer FNN keeping the number of neurons constant. Further the model is compared with other constructive methods like the Taylor's series implementation and Trapezoidal functions implementation using ResNet.  

\section{Motivation}
In digitone method of creating instrument sounds digitally, FM synthesis is used. Instead of using multiple frequencies to implementing the instruments , only a carrier and modulation frequency are used. In FM synthesis, double trigonometric functions are used. Hence, we are motivated to use double trigonometric functions which can be generated by double layer FOurier neural networks. Our idea is similar to digitone, here also we eould be able to approximate the target functions by using less frequencies, which imply less neurons in the Network.

We see that the feed forward network is able to approximate polynomial functions with less error. So they are ideal for implementing polynomial type of functions. The reason is that these networks follow a specific methodology which causes this. In this case, the target function is first approximated with Taylor series approximation function F1 which is in turn approximated with the Neural network. Thus, these Neural network can be said to be based on polynomial functions.

Similarly, the ResNet network we implemented is based on trapezoidal approximation. Consider a target function which is to be implemented. Its function graph is divided uniformly, with sample distance a. For every sample distance, the graph is approximated by a trapezoid.
The concatenation of all such trapezoidal functions,  approximates the target function. This this ResNet network can be considered to be based on trapezoidal function. We know that trapezoidal function in this case is set in a way to approximate the step function with step size a. Hence, this NN approximates well, the functions which are combinations of step functions. Or functions which can be well approximated by combination of step functions.

We implemented both networks  for different types of target functions. We observe that sinusoidal functions are not approximated well. We propose that this is because sinusoidal function by nature cannot be well approximated by polynomial functions or step functions. Hence we propose a Neural network which is based on sinusoidal functions. This new NN approximates the target functions by its Fourier series, i.e. combinations of sinusoidal functions. 
Hence it will be effective to approximate sinusoidal type of functions.

\section{Fourier Neural Network- Effectiveness}
Fourier Neural network as we predicted is able to approximate the sinusoidal functions effectively. But there are some problems. For implementing this NN, it is requiring so many neurons in the hidden layer. This is the main problem.
\section{Double Fourier Neural network}
To reduce the number of neurons required to approximate the target function , we propose Neural network that generate double trigonometric functions. Our motivation is based on the fact that deep Neural network in general approximates better. We find that in some cases, the network is giving good performance, while in others, its not very effective. We could not make solid conclusions based on the experiments.
\section{Hybrid Fourier Neural network}
We proposed a new hybrid Neural network which is combination of normal trigonometric functions and double trigonometric functions. This hybrid Neural network is applied to approximate the  target function by implementing corresponding Fourier series.  We conclude that this network is giving good results in most cases.

\section{Results And Discussion}\label{sec:results}
\begin{table*}[ht!]
\caption{Epsilon 1 using ResNet method}
\label{epsilon1 ResNet}
\renewcommand{\arraystretch}{1}
\renewcommand{\tabcolsep}{2.4mm}
\centering
\begin{tabular}{|l|l|l|l|l|l|}
\hline
\textbf{M}    & \textbf{sin(2*pi*x/5)} & \textbf{sin(2*pi*x/2.5)} & \multicolumn{1}{c|}{\textbf{\begin{tabular}[c]{@{}c@{}}rectfunc 1 to 10 \\ 1 cycle\end{tabular}}} & \multicolumn{1}{c|}{\textbf{\begin{tabular}[c]{@{}c@{}}rectfunc 1to10\\  2cycles\end{tabular}}} & \textbf{log(x)} \\ \hline
\textbf{5}    & 6.366198               & 6.366198                 & 0                                                                                                 & 2.5                                                                                             & 3.61822         \\ \hline
\textbf{10}   & 4.266574               & 7.531705                 & 0                                                                                                 & 1.111111                                                                                        & 2.250196        \\ \hline
\textbf{50}   & 0.814721               & 1.617057                 & 0                                                                                                 & 0                                                                                               & 0.898553        \\ \hline
\textbf{100}  & 0.403754               & 0.805831                 & 0                                                                                                 & 0                                                                                               & 0.519654        \\ \hline
\textbf{500}  & 0.080142               & 0.160234                 & 0                                                                                                 & 0                                                                                               & 0.112856        \\ \hline
\textbf{1000} & 0.040238               & 0.08008                  & 0                                                                                                 & 0                                                                                               & 0.031386        \\ \hline
\end{tabular}
\end{table*}

\begin{table*}[ht!]
\caption{Epsilon 2 using ResNet method}
\label{epsilon2 ResNet}
\renewcommand{\arraystretch}{1}
\renewcommand{\tabcolsep}{2.4mm}
\centering
\begin{tabular}{|l|l|l|l|l|l|}
\hline
\textbf{M}    & \textbf{sin(2*pi*x/5)} & \textbf{sin(2*pi*x/2.5)} & \textbf{rect\_1\_to\_10} & \textbf{rect\_1\_to\_10(2 cycles)} & \textbf{log(x)} \\ \hline
\textbf{5}    & 0.00056                & 0.00112                  & 0                        & 0                                  & 0.274802        \\ \hline
\textbf{10}   & 1.10E-13               & 1.86E-13                 & 0                        & 0                                  & 8.18E-12        \\ \hline
\textbf{50}   & 1.28E-11               & 1.57E-11                 & 0                        & 0                                  & 1.18E-10        \\ \hline
\end{tabular}
\end{table*}

\begin{table*}[ht!]

\caption{Epsilon 1 for Feedforward Networks}
\label{epsilon1 Srikant}
\renewcommand{\arraystretch}{1}
\renewcommand{\tabcolsep}{2.4mm}
\centering
\resizebox{175mm}{!}{%
\begin{tabular}{|c|c|c|c|c|c|c|c|c|c|}

\hline
\textbf{N}  & \multicolumn{1}{c|}{\textbf{gaussian}} & \multicolumn{1}{c|}{\textbf{x\textasciicircum{}2}} & \multicolumn{1}{c|}{\textbf{x\textasciicircum{}(-2)}} & \multicolumn{1}{c|}{\textbf{sinc2}} & \multicolumn{1}{c|}{\textbf{sin(2*pi*x/0.5)}} & \multicolumn{1}{c|}{\textbf{sin(2*pi*x/0.25)}} & \multicolumn{1}{c|}{\textbf{exp(x)}} & \multicolumn{1}{c|}{\textbf{exp(-x)}} & \multicolumn{1}{c|}{\textbf{sinc2\_new}} \\ \hline
\textbf{5}  & 0.006882                               & $5.55\times10^{-17}$                                           & 7.35E-08                                              & 1.86088                             & 2273.285                                      & 78403.85                                       & 2.22E-16                             & 2.22E-16                              & 2.263957                                 \\ \hline
\textbf{10} & 2.78E-16                               & 5.55E-17                                           & 9.86E+24                                              & 8.51E-13                            & 10781.7                                       & 3291133                                        & 0                                    & 2.05E-08                              & 2.22E-16                                 \\ \hline
\textbf{25} & 5.8E-15                                & 5.55E-17                                           & 5.46E-12                                              & 4.82E-14                            & 35.25349                                      & 3.04E+09                                       & 1.78E-15                             & 8.33E-16                              & 6.57E-08                                 \\ \hline
\textbf{50} & 2.78E-16                               & 5.55E-17                                           & 3.1E+105                                              & 4.54E+85                            & 6.26E-11                                      & 68970.15                                       & 1.78E-15                             & 8.33E-16                              & 2.94E-15                                 \\ \hline
\textbf{75} & 2.78E-16                               & 5.55E-17                                           & 5.46E-12                                              & 1.69E+86                            & 2.39E-11                                      & 5.44E-06                                       & 1.78E-15                             & 8.33E-16                              & 2.94E-15                                 \\ \hline
\end{tabular}%
}
\end{table*}

\begin{table*}[ht!]
\caption{Epsilon 2 for Feedforward Networks}
\label{epsilon2 Srikant}
\renewcommand{\arraystretch}{1}
\renewcommand{\tabcolsep}{2.4mm}
\centering
\resizebox{175mm}{!}{%
\begin{tabular}{|c|l|l|l|l|l|l|l|l|l|}
\hline
\multicolumn{1}{|l|}{\textbf{n}} & \multicolumn{1}{c|}{\textbf{gaussian}} & \multicolumn{1}{c|}{\textbf{x\textasciicircum{}2}} & \multicolumn{1}{c|}{\textbf{x\textasciicircum{}(-2)}} & \multicolumn{1}{c|}{\textbf{sinc2}} & \multicolumn{1}{c|}{\textbf{sin(2*pi*x/0.5)}} & \multicolumn{1}{c|}{\textbf{sin(2*pi*x/0.25)}} & \multicolumn{1}{c|}{\textbf{log(x)(from 0.1)}} & \multicolumn{1}{c|}{\textbf{exp(x)}} & \multicolumn{1}{c|}{\textbf{exp(-x)}} \\ \hline
\textbf{5}                       & 0.004827                               & 0.030599                                           & 6.701083                                              & 0.03125                             & 0.25                                          & 0.5                                            & 0.081827                                       & 0.053137                             & 0.01996                               \\ \hline
\textbf{10}                      & 0.000143                               & 0.000944                                           & 0.097095                                              & 0.000904                            & 0.007226                                      & 0.014422                                       & 0.002232                                       & 0.001552                             & 0.000567                              \\ \hline
\textbf{25}                      & 4.17E-09                               & 2.57E-08                                           & 2.88E-06                                              & 2.74E-08                            & 2.21E-07                                      & 4.4E-07                                        & 6.58E-08                                       & 4.73E-08                             & 1.78E-08                              \\ \hline
\textbf{50}                      & 1.28E-16                               & 8.13E-16                                           & 1.05E-13                                              & 7.96E-16                            & 6.31E-15                                      & 1.28E-14                                       & 2.21E-15                                       & 1.43E-15                             & 5.19E-16                              \\ \hline
\textbf{60}                      & 0                                      & 0                                                  & 0                                                     & 0                                   & 0                                             & 0                                              & 0                                              & 0                                    & 0                                     \\ \hline
\end{tabular}%
}
\end{table*}

\begin{table*}[ht!]
\caption{Error for Fourier Neural Networks}
\label{epsilon2 Srikant}
\renewcommand{\arraystretch}{1}
\renewcommand{\tabcolsep}{2.4mm}
\centering
\resizebox{175mm}{!}{%
\begin{tabular}{|l|l|l|l|l|l|l|l|l|l|}
\hline
\textbf{Function} & gaussian & x\textasciicircum{}2 & x\textasciicircum{}(-2) & sinc2    & sin(2*pi*x) & sin(4*pi*x) & exp(x)   & exp(-x)  & log(x)   \\ \hline
\textbf{Error}    & 9.66E-10 & 1.82E-08             & 4.9642                  & 1.26E-09 & 3.70E-04    & 1.05E-03    & 5.48E-05 & 5.39E-05 & 1.67E-03 \\ \hline
\end{tabular}%
}
\end{table*}

The tables 1,2 show the error values of ResNet architecture. As discussed before, epsilon 1 here, is error  between ResNet and the concaetination of rectangular functions . We take M samples of functions and construct the rectangular functions with the sample height for the sample duration. In the table 1, rectfunc 1 to 10 (1 cycle) means function is rectangular function with output 1 from input values in the range 0 to 5 and outputs -1 for input values in the range 5 to 10. Whereas the function in table shown as rectfunc 1 to 10 (2 cycles) means function is rectangular function with output  1 when input is in range 0 to 2.5. Output is -1 when input is in range 2.5 to 5, again output is 1 for input in range 5 to 7.5 and again output is -1 for input in range 7.5 to 10. Thus this function has 2 cycles between 0 and 1. 
It can be observed that For Table 1, error is least for rectangular type of functions which is expected. In the case of sine function approximation, it is found thatfor sinsoidal function of higher frequency, the error is more.The log(x) function is considered from 0.01 to 10. 
In table 2, it is observed that error is greatly decreased for M=50. It is also noticed that for epsilon2 error in Table 2, the error is almost independent of the nature of input function , for sufficiently large number of samples(greater than 50) and the error is less than $10^{-10}$.

In case of feed forward network implementation, the results are given in Table 3 and Table 4. In table 3, we observe that error converges in all functions except $x^{-2}$ and $sinc^{2}$ functions. Further when comparing the sinusoidal functions of diiferent frequencies, the function with high frequency gets less error. Error in this table are direct consequence of Remainder theorem in Taylor series. It is clear that error is least for polynomial functions of positive integer degrees. $sinc^2_new$ indicates the taylor series is calculated for sinx and then divided by x. For functions $sinc^2$ and exp(-x) , taylor  series is taken around 0.01, while for all other values taylor series is calculated around 0.

The Table 4, gives errors between the Neural network and taylor series implementation. It is observed that for sufficient number of terms in binary expansion of x given by 'n', error is almost zero. 

Table 5 , gives the error between double layer Fourier Neural network which is earlier refered as Hybrid Fourier Neural Network. This experiment is conducted taking 10000 samples of x in [-1,1]. It is observed that the error is more for $x^{-2}$ since function becomes undefined at x=0. Hence samples around x=0 give large values leading to error. For all other functions, error is fairly contained. For all the experiments in Tables 1-5, error is 2-norm error.

\section{Conclusion}
The deficiencies of the neural network as universal approximator proposed in the literature are explored and discussed. Some of the existing architectures are explored which are found to be relatively inefficient when approximating sinusoidal like target functions. Taking hint from the previous architectures, a FNN architecture is proposed to tackle such target functions. We extend this further by considering FNN with two hidden layers. It is proven that such a FNN output cannot be expressed in Fourier series and hence it cannot be proved as universal approximator unlike single hidden layer FNN which is a Universal Approximator. A novel architecture called "modified Architecture" is proposed which is proven to be more effective than FNN. Using two hidden layers, it is seen that number of neurons required is less compared to Single layer FNN.

\bibliographystyle{IEEEbib}
\bibliography{refs}

\end{document}